\documentclass[
  journal=medium,
  manuscript=article-type,
  year=2023,
]{cup-journal}

\usepackage{amsmath}
\usepackage[nopatch]{microtype}
\usepackage{booktabs}

\title{Beyond Sentiment: Leveraging Topic Metrics for Political Stance Classification}

\author{Weihong Qi}
\affiliation{University of Rochester}
\email[Weihong Qi]{wqi3@ur.rochester.edu}

 %% First letter not capped

\begin{document}

\begin{abstract}

Sentiment analysis, widely critiqued for capturing merely the overall tone of a corpus, falls short in accurately reflecting the latent structures and political stances within texts. This study introduces topic metrics, dummy variables converted from extracted topics, as both an alternative and complement to sentiment metrics in stance classification. By employing three  datasets identified by \citet{bestvater2023sentiment}, this study  demonstrates BERTopic's proficiency in extracting coherent topics and the effectiveness of topic metrics in stance classification. The experiment results show that BERTopic improves coherence scores by 17.07\% to 54.20\% when compared to traditional approaches such as Dirichlet Allocation (LDA) and Non-negative Matrix Factorization (NMF), prevalent in earlier political science research. Additionally, our results indicate topic metrics outperform sentiment metrics in stance classification, increasing performance by as much as 18.95\%. Our findings suggest topic metrics are especially effective for context-rich texts and corpus where stance and sentiment correlations are weak. The combination of sentiment and topic metrics achieve an optimal performance in most of the scenarios and can further address the limitations of relying solely on sentiment as well as the low coherence score of topic metrics.

\end{abstract}

\section{Introduction}
Sentiment analysis has emerged as an increasingly prominent method in political text analysis. Numerous studies have highlighted its ability to extract valuable insights from political texts, such as public sentiment in general election cycle \citep{wang2012system}, capturing the political and economic narratives \citep{ash2023relatio}, emotions toward major political parties \citep{ansari2020analysis}, and attitudes toward controversial policies like COVID-19 vaccinations policy \citep{lyu2022social}. However, sentiment analysis has been critiqued for merely capturing the overall tone of a text, failing to adequately represent the specific political stance that is targeted and pertains to individual topics or interest entities \citep{bestvater2023sentiment}. Furthermore, existing research emphasizes that the effectiveness of political stance classifiers is not uniform and is influenced by the type and quality of the text data \citep{grimmer2013text, gonzalez2015signals}. As scholars have acknowledged the distinctions between sentiment and stance, and recognized the limitations of sentiment analysis, many studies have attempted to enhance the stance classification techniques by using target-based methods \citep{kuccuk2020stance}. Others have explored alternatives to sentiment in textual features, such as n-grams \citep{elfardy2016cu}. However, despite the advancement of topic modeling methods, to the best of our knowledge, no research has leveraged the capacity of innovative topic modeling methods in identifying the latent structures within texts in political stance classification.

In this article, we explore the potentials of topic metrics, \textit{which are dummy variables converted from extracted topics}, to be an alternative solution for stance classification when sentiment analysis falls short in the task. Stance is defined as \textit{the emotional or attitudinal position expressed toward a specific target and is often linked to distinct topics or entities of interest} \citep{mohammad2016semeval, bestvater2023sentiment}. In addition, recent media studies indicate the potentials of topic metrics in capturing diverse targets influenced by social groups and political ideologies, finding that political stances are reflected in variations of fine-grained themes \citep{panbias}. Therefore, topic metrics can potentially address the shortcomings of sentiment metrics in stance classification.

To investigate the potentials of topic metrics in political stance classification, we study the following research questions in this article:
\begin{itemize}
\item RQ1: What is the most effective topic modeling method for analyzing political texts?
\item RQ2.1: Can topic metrics outperform sentiment metrics in classifying stance?
\item RQ2.2: Does the combination of topic metrics and sentiment better than only using one of them?

\item RQ3: What criteria should inform the choice of using either of the two metrics, or a combination of both, for classifying political stances?

\end{itemize}

Using three dataset with various types and contexts as identified by \citet{bestvater2023sentiment}, we first illustrate that BERTopic outperforms traditional topic modeling techniques, including Latent Dirichlet Allocation (LDA) and Non-negative Matrix Factorization (NMF), in achieving topic coherence. We then provide evidence that the topics extracted through BERTopic, serving as topic metrics, surpass sentiment metrics in classifying political stances, particularly when analyzing context-rich texts from social media platforms. The topic metrics can improve the performance in stance classification over sentiment by 18.95\% when applied to the context-rich texts from social media platforms. The combination of the topic and sentiment metrics show the best performance when sentiment has certain level of correlation with stance. However, its performance diminishes and topic metrics perform optimally when the correlation is weak.

Our results suggest that BERTopic generate more coherent topics than LDA and NMF in political text analysis. Topic metrics are best suited for context-rich texts from social media and surpass sentiment metrics in other text types. However, combining both metrics into political stance classification does not consistently improve model performance, as a weak correlation between sentiment and stance might introduce noise during model training. When the coherence score is low, the combination of the two metrics offers the most significant enhancement in performance compared to using topic metrics alone. Thus, for stance classification in social media or survey response texts, topic metrics should be prioritized over sentiment. If there is a noticeable correlation between sentiment and stance, or if the coherence score is low, the combination of both metrics is optimal.

The topic metrics, demonstrated as a viable alternative or supplement to sentiment metrics, reveal multiple advantages. Firstly, they minimize the number of assumptions and biases, and the labor required compared to manual political stance labeling, while enhancing accuracy over sole reliance on sentiment metrics. Secondly, they are extracted from current data as a new measurement and there is no need for additional data collection. Lastly, they provide political scientists insights into which issues present greater topic variances in debates versus those dominated by sentiments.

This article primarily contributes in introducing an innovative metric for the classification of political stances, particularly within the context-rich texts from social media. A secondary, yet equally significant contribution is highlighting the need for future political science research to consider topic discrepancies, along with their associated framing and attention differentials, as vital components in capturing the political stance conveyed in political texts. Furthermore, this article shows a novel application of topic modeling methods in the realm of political text analysis.

\newpage

\section{Methodology}
\subsection{Topic Modeling Approach}
Topic modeling is an unsupervised machine learning technique employed across a wide range of natural language processing (NLP) tasks, including political text analysis \citep{ying2022topics, jelodar2019latent, roberts2014structural}. While it is primarily used in text mining to extract abstract topics from a corpus, it also reveals the latent structures within texts without necessitating prior knowledge about the corpus \citep{shi2018short, chauhan2021topic, grootendorst2022bertopic}. In the realm of topic modeling, traditional methodologies are predominantly characterized by LDA \citep{bagozzi2018politics, wilkerson2017large} and NMF \citep{greene2017exploring}, representing probabilistic and matrix-based approaches, respectively. Nevertheless, with the advent of pre-trained large language models (LLMs), recent findings indicate that the transformer-based model, BERTopic, surpasses both LDA and NMF in terms of topic modeling coherence, especially in the context of social media text analysis \citep{egger2022topic}.

BERTopic is derived from the pre-trained language model, Bidirectional Encoder Representations from Transformers (BERT) \citep{devlin2018bert}. While LDA and NMF employ the bag-of-words method to represent text documents, BERTopic leverages contextual word and sentence vector representations. Through the integration of clustering techniques and a class-centric adaptation of Term Frequency - Inverse Document Frequency (TF-IDF) method, BERTopic formulates more coherent topic representations. This indicates that BERTopic could perform better in context-rich tasks. Empirical studies find that BERTopic consistently outperforms LDA and NMF in a multitude of topic modeling tasks \citep{grootendorst2022bertopic}.

BERTopic-generated topics are potential alternatives for political stance classification when the relation between stance and sentiment is weak. Such a scenario arises when discernible differences in sentiments are weak, suggesting that approving and opposing stances may focus on distinct facets of an issue. For instance, in the U.S. abortion debates, pro-abortion Twitter users frequently emphasize constitutional rights and freedoms for women, represented by hashtags such as \textit{\#prochoice}; conversely, anti-abortion users rally by \textit{\#prolife} \citep{sharma2017analyzing}, showing the intention to ``protect and nurture human life at every stage of its existence'' \citep{prolife}. 

The study by \citet{bestvater2023sentiment} provides examples that further demonstrate this observation. In their research on tweets regarding Kavanaugh's confirmation, the authors identified a weak correlation between stance and sentiment. A closer examination of the dataset reveals that approving and opposing stances highlight different aspects of the topic, while they both exhibit similar sentiment. For example, some tweets in approving stance, as illustrated in the following cases, emphasize the Democrats' lack of evidence and express empathy toward Kavanaugh and his family, expressing negative sentiment:

\begin{quote}
Approving tweet (1): ``When nothing is proved by this Ford nut case Judge Kavanaugh needs to file a law suit for defamation of character. For the lies and pain caused to him and his family. These bottom feeders ( Democrats.) need to be taught a lesson and Feinstein should be removed from her seat!!! ''
\end{quote}

\begin{quote}
Approving tweet (2): ``I'm more concerned about how Judge Kavanaugh's wife and daughters feel than how Christine Ford feels.  She's feeling "anxiety"?  I would too if I was about to perjure myself in front of the whole planet.''
\end{quote}

Conversely, the opposing tweets question Kavanaugh's professionalism as a judge and highlight his alleged past misconduct, but also written in negative sentiment:

\begin{quote}
Opposing tweet (1): ``For a guy who’s a judge, Brett Kavanaugh seems to really hate the idea of going through a fair, methodical process where information is gathered and opposing points of view are expressed so that we can find out the truth.''
\end{quote}

\begin{quote}
Opposing tweet (2): ``Putting aside everything else for a moment, Kavanaugh's murky financials are major, major red flags. Dark money. Unexplained debt payments. There is very clearly a concerted, monied effort running in the background.''
\end{quote}

Apart from aligning with the inherent nuances of political stance division, BERTopic also exhibits computational efficiency as it does not require additional computational capacity such as GPUs, which are common in classification methods like fine-tuning pre-trained large language models \citep{dodge2020fine, wang2023topic}. Furthermore, it offers advantages in terms of resource conservation relative to crowd-sourcing and introduces fewer assumptions than human-labeling approaches.

\subsection{Data Pre-processing}
In this article, we utilize three datasets as identified by \citet{bestvater2023sentiment}. These datasets encompass both survey and social media data across diverse contextual lengths. Specifically, the first dataset comprises tweets related to the 2017 Women's March (WM). The second dataset pertains to tweets about the Kavanaugh Confirmation (KC).  The last dataset draws from open-ended responses in the Mood of the Nation Poll (MOTN) concerning the approval of Donald Trump. Among the three datasets, KC features the longest texts, with an average length of 16.8 tokens, while WM contains the shortest texts, averaging 7.2 tokens in length and MOTN has the average number of tokens to be 13.1.

Prior to the data analysis, we remove the numbers, punctuation and emojis in the corpus and conduct lemmatization. We also use the English dictionary built in the Natural Language Toolkit (NLTK) library to remove the stop words.\footnote{\url{https://www.nltk.org/}} The generated topics are converted to dummy variables with One-Hot Encoder.\footnote{\url{https://scikit-learn.org/stable/modules/generated/sklearn.preprocessing.OneHotEncoder.html}}

\subsection{Model Evaluation}

In this research, we first conduct an experiment on the three datasets, fine-tuning hyperparameters and determining the most coherent topic modeling method based on coherence scores. The coherence score, which estimates the semantic coherence of the topics and the texts, reflects the level of interpretability of the generated topics \citep{syed2017full}. Then we compare the F1 scores obtained from sentiment metrics, topic metrics and the combination of the two metrics for stance classification on the three datasets. We report the results using Logistic model as it is a rudimentary classification method and the performance of models using topic metrics could find improvement with more complex classifiers. The results with other classification models, including K-Nearest Neighbor (KNN), Support Vector Machine (SVM) and XGBoost are reported in the Appendix. To train the classification models, we split each data with 80\% of the sample to be the training set and the remaining 20\% to be the test set. The classification experiments are conducted with 10-fold cross validations to avoid the over-fitting problems. The outcomes of this experiment can guide the choice of sentiment, topic and the combined metrics in political stance classification.

\section{Results}
\subsection{Topic Extraction}

Figure 1 illustrates the coherence scores obtained when employing LDA, NMF, and BERTopic on various datasets, adjusting hyperparameters and varying the number of topics from 5 to 50. Across all three datasets, topics generated by BERTopic consistently secured the highest coherence scores, confirming BERTopic's robust interpretability. Table 4 in the Appendix presents the topics generated by BERTopic when the number of topics is set to 10. Following standard practices in research that uses topic modeling methods, we manually assign topic labels based on extracted keywords \citep{okon2020natural, wu2021characterizing}. The extracted topics reveal informative keywords that are distinct across topics and easy to label, further confirming the strong interpretability indicated by the coherence scores.

Meanwhile, it is also essential to notice the subtle variations in model performance within individual datasets. For the WM and MOTN dataset, BERTopic's optimal performance is realized with a topic count of 15. In contrast, for the KC dataset, which is characterized by a richer context, the model's efficiency continually ascends with an increment in topic count. Notably, as shown in Table 1, the superiority of BERTopic over traditional topic modeling methods is most significant for the KC dataset, making an enhancement of 54.20\%. This increment is markedly greater than the performance improvement observed when BERTopic is applied to the other two datasets, which recorded enhancements of 19.12\% and 17.08\%, respectively. 

\textbf{The findings indicate that BERTopic consistently delivers better topic coherence performance across diverse data types and contexts. Furthermore, BERTopic's coherence enhancement is most evident in datasets with a richer context.}

\begin{figure}[hbt!]
\centering
\includegraphics[width=\linewidth]{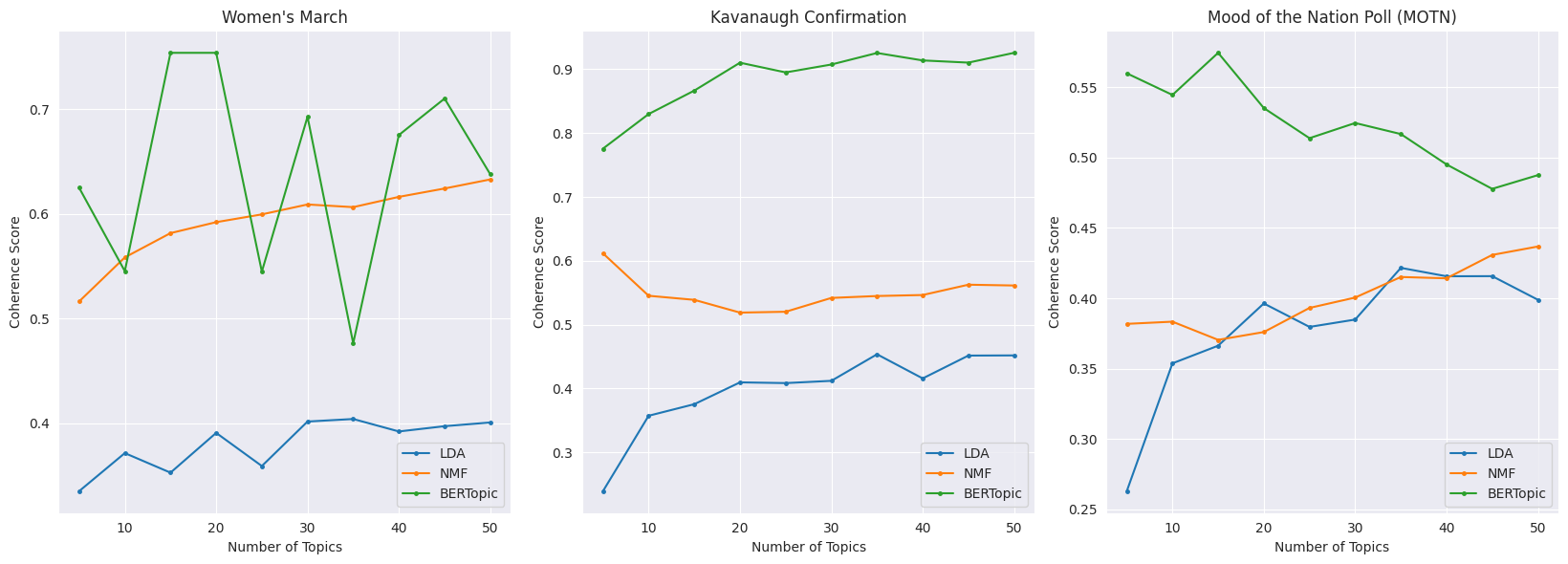}
\caption{The Coherence Scores with Varying Hyperparameters}
\label{}
\end{figure}

\begin{table}[hbt!]
\begin{threeparttable}
\caption{The Best Coherence Scores Achieved by LDA, NMF and BERTopic on Different Dataset}
\label{table_example}
\begin{tabular}{l*{6}{c}}
\toprule
\headrow Dataset &  \multicolumn{3}{c}{Coherence Score} & Context & Data Type & Evaluation \\
  & LDA & NMF & BERTopic & Avg. Tokens & Type & BERTopic enhancement  \\ 
\midrule
WM & 0.4006 & 0.6329 & \textbf{0.7539} & 7.2 & social media & 19.12\% \\ 
KC & 0.4518 & 0.6116 & \textbf{0.9431} & 16.8 & social media & 54.20\% \\
MOTN & 0.4216 & 0.4367 & \textbf{0.5113} & 13.1 & survey data & 17.08\% \\
\bottomrule
\end{tabular}

\end{threeparttable}
\end{table}

\subsection{Stance Classification}
Building on the findings from Section 3.1, we implement an experiment to assess the performance of topics, generated by BERTopic from three datasets, as alternate or supplement metrics to sentiment for political stance classification. We present the results using the Logistic model as the classifier in Table 2, while the outcomes from KNN, SVM, and XGBoost are reported in Table 5 in the Appendix.

Table 2 shows the performance of topic, sentiment and the combination of the two metrics, with the Logistic model as the classifier. 
Overall, topic metrics outperform sentiment metrics on all three datasets when assessed by the F1 score. Combining the two metrics yields optimal results on WM and MOTN. However, on the KC dataset, where the correlation between sentiment and stance is notably weak, using only the topic metrics achives the highest F1 score.

Specifically, for the WM dataset, topic metrics predict political stance with an F1 score of 0.9347, slightly up from 0.9281 achieved with sentiment metrics. A combination of both metrics surpasses either metric alone, attaining an F1 score of 0.9366. Similarly, for the MOTN dataset, topic metrics outperform sentiment metrics, with the combined approach elevating the F1 score from 0.5705 (using only sentiment) to 0.6516. For the KC dataset, the topic metrics significantly outperform the sentiment metrics and also achieve a slightly higher F1 score than the combined metrics. Results using other classifiers, as reported in Table 5 in the Appendix, follow patterns similar to the Logistic model without showing significant enhancement. However, when using KNN, SVM, and XGBoost on the KC dataset, the combined metrics surpass topic metrics alone. This suggests that more complex models may overcome the noise from the weak correlation between sentiment and stance.

To provide insights into the criteria guiding metric selection, we closely study the relationship between performance enhancement and dataset characteristics. Table 3 presents the relationship between the improvement over sentiment and both topic coherence and the correlation between sentiment and stance. It is noteworthy that the most significant improvement is observed on the KC dataset, where topic and combined metrics enhance performance over sentiment by 18.95\% and 18.68\%, respectively. Given that KC has the most extended context, and considering BERTopic's proficiency in within-context learning, coupled with sentiment's inadequacy in capturing stance, this suggests that \textbf{topic and combined metrics are ideal for context-rich texts and corpus where there's a weak correlation between stance and sentiment}. For MOTN, topic and combined metrics exhibit improvements of 5.7\% and 14.22\%, respectively. \textbf{This suggests that combined metrics can address the limitations of sentiment in stance classification, as well as the low coherence score of topic metrics.} The enhancement for the WM dataset is minimal, under 1\%. This can be attributed to both sentiment and topic metrics already reaching a notably high F1 score, exceeding 0.92.

\begin{table}[hbt!]
\begin{threeparttable}
\caption{The Performance of topic, Sentiment, and Combined Metrics in Stance Classification}
\label{table_example}
\begin{tabular}{l*{5}{c}}
\toprule
\headrow Dataset &  \multicolumn{3}{c}{F1 Score} & Correlation \\
  & Topic & Sentiment & Topic and Sentiment& $Corr(Stance, Sentiment)$ \\ 
\midrule
WM & 0.9347 & 0.9281 &  \textbf{0.9366} & 0.44 \\ 
 & (0.0027) & (0.0002) &  (0.0021) & \\
KC & \textbf{0.8373} & 0.7039 & 0.8354 & 0.03  \\
 & (0.0217) & (0.0006) &  (0.0234) & \\
MOTN & 0.6030 & 0.5705 & \textbf{0.6516} & 0.51  \\
 & (0.0342) & (0.0180) & (0.0280) & \\
\bottomrule
\end{tabular}
\end{threeparttable}
\end{table}

%When analyzing the characteristics of the two datasets where attention metrics and sentiment display the most divergent performance, a key observation emerges: the correlation between sentiment and stance stands at 0.03 for the KC dataset, whereas it is 0.51 for the MOTN dataset. Additionally, the KC dataset exhibits greater context complexity compared to both WM and MOTN datasets. Taking into account the dataset features and the performance differences when applying the two metrics for political stance classification, one can deduce that \textbf{topic metrics excel in stance classification particularly when the dataset is context-rich, sourced from social media, and exhibits a tenuous correlation between sentiment and stance.}

\begin{table}[hbt!]
\begin{threeparttable}
\caption{
F1 Improvement, Topic Coherence and Sentiment-Stance Correlation.}
\label{table_example}
\begin{tabular}{l*{5}{c}}
\toprule
\headrow Dataset &  \multicolumn{2}{c}{F1 Improvement over Sentiment} & Coherence & Correlation \\
 & Topic &  Combination  & Topic Coherence Score& $Corr(Stance, Sentiment)$ \\ 
\midrule
WM & 0.71\%  & 0.92\%  & 0.7539 & 0.44  \\ 
KC & 18.95\% & 18.68\% & 0.9431 & 0.03 \\
MOTN & 5.70\% & 14.22\% & 0.5113 & 0.51  \\
\bottomrule
\end{tabular}
\end{threeparttable}
\end{table}

\section{Conclusion}
Sentiment analysis, as critiqued by \citet{bestvater2023sentiment}, often captures merely the general tone of a corpus, making it less effective in political stance classification. In this study, we introduce an innovative metric to be an alternative and complement for this task. Our experiment results provide evidence showing that topic metrics, derived from topic modeling methods, can outperform sentiment metrics in classifying stance. Combining both metrics yields optimal results in most scenarios. Topic metrics extracted via BERTopic, which achieve significantly better coherence scores compared to traditional LDA and NMF, can enhance political stance classification by up to 18.95\% over sentiment metrics alone. Our results further suggest that topic and combined metrics are particularly suited for context-rich texts with a weak link between stance and sentiment. Additionally, the combined metrics can overcome sentiment's limitations in stance classification and the low coherence of topic metrics.

The findings of the paper not only provide an alternative and complement metrics for political stance classification, but also offers insights for future research to pay more attention to the topic discrepancies in political texts about controversial topics. Methodologically, future research can extend the application of the topic metrics to other types of texts, especially longer texts such as speeches and official documents.

%we put forth a novel metric for discerning political stance. This attention metric, defined by topics generated through BERTopic, discerns varied emphases on different facets of an issue across diverse groups. Particularly when the linkage between sentiment and stance is tenuous, this metric aids in classifying political stance. Our findings indicate that BERTopic outperforms LDA and NMF in generating coherent topics, especially for context-rich data from social media. Additionally, we furnish evidence that topics generated by BERTopic, functioning as attention metrics, surpass sentiment in stance classification, most notably for WM and KC datasets—both derived from social media. The augmentation in performance is most salient with context-rich data. These results proffer a guideline for opting between attention and sentiment metrics: the former is apt for context-rich social media datasets with a feeble sentiment-stance correlation, whereas the latter is optimal for succinct survey texts.

%\begin{acknowledgement}
 %The author thanks Professor Jiebo Luo, 
%\end{acknowledgement}

%\endnote in some journals will behave like \footnote; and \printendnotes will not output anything. 
%\printendnotes

\printbibliography

\newpage

\appendix
\section{Appendix}

\begin{table}[hbt!]
\centering
\resizebox{\columnwidth}{!}{%
\begin{threeparttable}
\caption{Top 10 Topics Generated by BERTopic and Associated Keywords}
\label{table_example}
\begin{tabular}{l*{2}{c}}
\toprule
\headrow \multicolumn{2}{c}{Women's March} \\
\midrule
Topics & Keywords \\
\midrule
Women's March Participation & March, Women, Today, Men, Everyone, World, Proud \\
Trump Administration Critique & \#realdonaldtrump, POTUS, President, \#seanspicer, Trump, Russia, Putin, Say, CIA \\
Signature & Sign, Favorite, Best, Some, My, Carry, Saw, One, McKellen \\
March Advocacy & Something, \#womensmarchlondon, Let, Yes, \#womensmarch, \#addhername, Thread, Get, Sly, \#vincentdonofrio \\
Washington March Events & Washington, DC, Columbia, District, Live, Capitol, via, \#mayorbowser, Watch, Monument \\
Peaceful Protests & Protest, Protester, Peaceful, Something, Dress, Goal, Nationalism, Democracy\\
Honoring Hillary Clinton & \#hillaryclinton, Hillary, \#addhername, Clinton, Honor, List, Honoree, She, Name, Quote \\
March Photos & Photo, Picture, Pic, Image, Some, See, Post, Favorite, Collection \\
Inauguration & Inauguration, \#inaugurationday, Crowd, Trump, Big, Attendance, Size, Attend, People, Press \\
Gender Equality & Equality, Worker, Pay, Sex, \#equalityforall, Right, Men, Wage, Peace
\\ 
\headrow \multicolumn{2}{c}{Kavanaugh Confirmation} \\
\midrule
Topics &  Keywords  \\
\midrule
Reputation and Political Implications & Wing, Facts, Destroy, Matter, Fine, Answer, Reputation, Politician, \#realdonaldtrump, Single \\
Misjudgment and Delays & Misread, Delays, Blowback, Sentiment, Grossly, You, Serious, Accommodate, Regard, Fail \\
Pundit Opinions and Criticisms & \#newrules, Loon, Pundits, Blogger, Head, Snot, Living, Sue, Basic, Either \\
Political Allegations & Guilty, Obama, Innocent, Kenneth, English, Crony, Adriandt, Karma, Treason, Asharangappa \\
Ford's Testimony & Recall, Ford, Ally, Dr, Happen, None, Anyway, Attempt, Nothing, Lesssavory \\
Allegations against Kavanaugh & Rape, Attempted, Cover, Graham, Crime, White, Lifetime, Fox, Prepared, Executive \\
Support to Kavanaugh & Proud, Colleague, Stand, Female, Jaymaga, Caring, Recounts, Gift, \#standwithkavanaugh, Friend \\
Impeachment & Impeach, Float, Impeachment, \#foxnews, Even, Democrats, Confirm, Barefoot, Bare, Nutjob \\
Anita Hill's Hearing & Ineptitude, Volcanic, Officer, State, Anger, Become, Derail, Hill, Lead, Keith \\
Emotions on the Hearing & Amusing, Sleepy, Piss, Lessen, Clown, Partisan, Chance, Awake, November, Still \\

\headrow \multicolumn{2}{c}{Mood of the Nation Poll} \\
\midrule
Topics & Keywords \\
\midrule
Trump's Reelection & Trump, Donald, Reelecting, Face, Every, General, Refineries, Nutty, Manboy, Believe \\
Accomplishments & President, Donald, Trump, Accomplishments, King, Word, Always, Aggressive, Harnessing, Offense \\
Democratic Obstruction & Democrats, They, Trying, Do, Blocking, Everything, Don't, to \\
War Concerns & War, US, into, Start, Wrold, Get, Another, Going, Starting, or \\
MAGA & Great, Again, Make, America, is, Continue, We, Think, to \\
Media Bias & Media, Mainstream, They, Left, Bias, Press, Attacks, the, Constant \\
Dishonesty and Lies & Lie, Truth, Cheating, Tells, Liar, Doesn't, and, the \\
Protests and Violence & Protests, Protesters, against, Riots, Rallies, Violent, Sore, Demonstration \\
Involvement & Anything, Related, Everything, Do, about, Involving, Literally, Relating, Basically\\
Nationalism & Proud, Me, Makes, American, Pride, Feel, is, Has, of \\

\bottomrule
\end{tabular}

\end{threeparttable}%
}
\end{table}

\begin{table}[hbt!]
\begin{threeparttable}
\caption{Metric Performance with KNN, SVM and XGBoost Models}
\resizebox{\columnwidth}{!}{%
\begin{tabular}{l*{10}{c}}
\toprule
\headrow Dataset &  \multicolumn{3}{c}{KNN} &  \multicolumn{3}{c}{SVM} &  \multicolumn{3}{c}{XGBoost} \\
  & Topic & Sentiment  & Topic and Sentiment & Topic & Sentiment  & Topic and Sentiment &  Topic & Sentiment  & Topic and Sentiment  \\ 
\midrule
WM & 0.9070 & 0.9075 & \textbf{0.9191} &  0.9348 & 0.9281 & \textbf{0.9377} & 0.9349 & 0.9281 & 0.9375\\ 
 & (0.0074) & (0.0314)  &  (0.0063) &  (0.0027) & (0.0002) & (0.0032) & (0.0028) & (0.0002) & (0.0029)\\
KC & 0.8163 & 0.4745 & \textbf{0.8292} & 0.8345 & 0.7039  & \textbf{0.8501} & 0.8335 & 0.7039 & \textbf{0.8343} \\
 & (0.0253)  & (0.3119) & (0.0219) & (0.0236) & (0.0006)  & (0.0162) & (0.0166) & (0.0006) & (0.0173) \\
MOTN & 0.6030 & 0.3511  & \textbf{0.6482} & 0.5795 & 0.5705  & \textbf{0.6864} & 0.5546 & 0.5705 & 0.6758\\
 & (0.0342) & (0.2872) & (0.0195)  & (0.0307) & (0.1080) & (0.0205) & (0.0191) & (0.1080) & (0.0214) \\
\bottomrule
\end{tabular}
}
\end{threeparttable}
\end{table}

\end{document}